\documentclass{article}

\usepackage{takeuchi_mcr}  
\usepackage{bm}
\usepackage{microtype}
\usepackage{graphicx}
\usepackage{subfigure}
\usepackage{booktabs}    
\usepackage{hyperref}
\usepackage{amsmath}
\usepackage{amssymb}
\usepackage{mathtools}
\usepackage{amsthm}
\usepackage{multirow}
\usepackage{makecell}
\usepackage{booktabs}
\usepackage[numbers]{natbib}
\usepackage{algpseudocode}
\usepackage{algorithm}
\usepackage{slantsc}
\usepackage[textsize=tiny]{todonotes}
\usepackage{authblk}
\usepackage{rotating}

\theoremstyle{plain}
\newtheorem{theorem}{Theorem}[section]

\newtheorem{lemma}[theorem]{Lemma}

\theoremstyle{definition}
\newtheorem{definition}[theorem]{Definition}

\theoremstyle{remark}

\title{Generalized Kernel Inducing Points by Duality Gap for Dataset Distillation}

\author[1]{Tatsuya Aoyama\thanks{Equal contribution.}}
\author[1]{Hanting Yang\thanks{Equal contribution.}}
\author[2]{Hiroyuki Hanada}
\author[1]{Satoshi Akahane}
\author[1]{Tomonari Tanaka}
\author[1]{Yoshito Okura}
\author[3]{Yu Inatsu}
\author[2]{Noriaki Hashimoto}
\author[4]{Taro Murayama}
\author[4]{Hanju Lee}
\author[4]{Shinya Kojima}
\author[1,2]{Ichiro Takeuchi}

\affil[1]{Nagoya University, Nagoya, Japan}
\affil[2]{RIKEN, Nagoya, Japan}
\affil[3]{Nagoya Institute of Technology, Nagoya, Japan}
\affil[4]{DENSO CORPORATION, Kariya, Japan}
\affil[]{{\tt takeuchi.ichiro.n6@f.mail.nagoya-u.ac.jp}}

\begin{document}
\maketitle

\begin{abstract}
We propose \textit{Duality Gap KIP} (DGKIP), an extension of the Kernel Inducing Points (KIP) method for dataset distillation. While existing dataset distillation methods often rely on bi-level optimization, DGKIP eliminates the need for such optimization by leveraging duality theory in convex programming. The KIP method has been introduced as a way to avoid bi-level optimization; however, it is limited to the squared loss and does not support other loss functions (e.g., cross-entropy or hinge loss) that are more suitable for classification tasks. DGKIP addresses this limitation by exploiting an upper bound on parameter changes after dataset distillation using the duality gap, enabling its application to a wider range of loss functions. We also characterize theoretical properties of DGKIP by providing upper bounds on the test error and prediction consistency after dataset distillation. Experimental results on standard benchmarks such as MNIST and CIFAR-10 demonstrate that DGKIP retains the efficiency of KIP while offering broader applicability and robust performance.
\end{abstract}


\section{Introduction}
\label{Introduction}

Reducing the amount of training data while preserving model performance remains a fundamental challenge in machine learning. \emph{Dataset distillation} seeks to generate synthetic instances that encapsulate the essential information of the original data \cite{yu2023dataset}. This synthetic approach often proves more flexible and can potentially achieve greater data reduction than simply retaining subsets of actual instances. Such distilled datasets can also serve broader applications, for example by enabling efficient continual learning with reduced storage demands \cite{masarczyk2020reducing, rosasco2021distilled, enneng2023efficient}, and offering privacy safeguards through data corruption \cite{dong2022privacy,liu2023backdoor}.

\begin{figure}[t]
\vskip 0.2in
\centering
\includegraphics[width=\columnwidth]{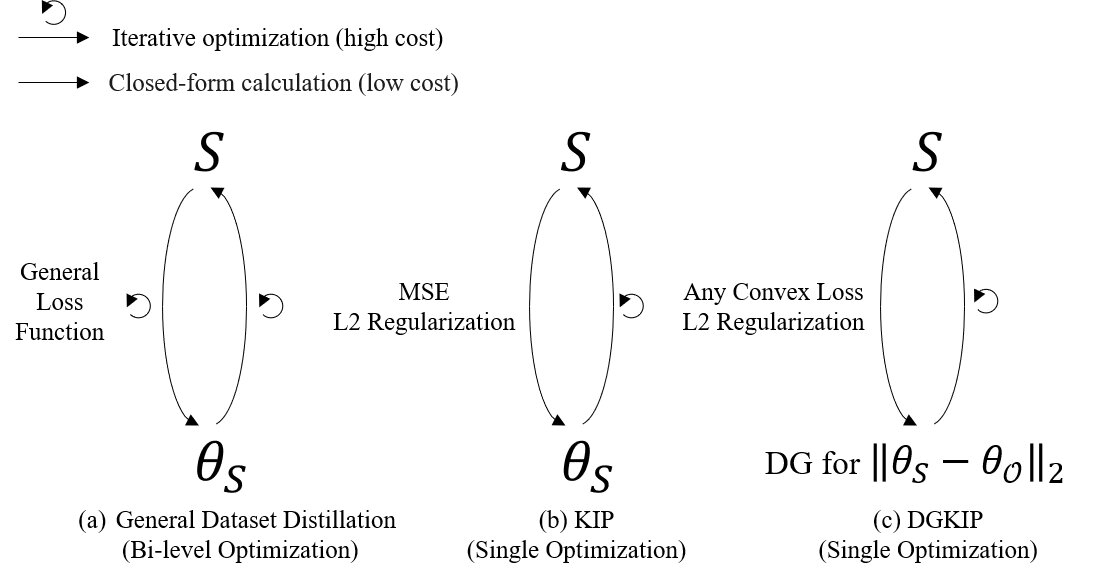}
\caption{Paradigm of dataset distillation with (a) Bi-level Optimization, (b) Kernel Inducing Points (KIP), and (c) proposed DGKIP. In each subfigure, $S$ means the synthetic data, $\theta_S$ means the model trained on $S$, and the arrow represents the optimization process. KIP method avoids bi-level optimization by simplified inner loop but restricted to square error loss, while DGKIP expands its paradigm to a large class of loss functions by evaluating the duality gap (DG) instead of $\bm\theta_\mathcal{S}$ itself.}
\label{solutionspace}
\vskip -0.2in
\end{figure}

Existing dataset distillation methods are essentially formulated as a bi-level optimization problem. This is because generating synthetic instances requires retraining the model with those instances as training data. Specifically, synthetic instances are created in the outer loop, and the model is trained in the inner loop, leading to high computational costs. A promising approach to avoid bi-level optimization is a method called \emph{Kernel Inducing Point (KIP)} \cite{nguyen2020dataset}. The KIP method avoids bi-level optimization by obtaining an analytical solution in its inner loop, effectively leveraging the fact that its loss function is a variant of squared loss. Although the KIP method is a type of kernel method, it can also be applied to dataset distillation in deep learning models by utilizing Neural Tangent Kernel (NTK) \cite{novak2019neural} or Neural-Network Gaussian Process (NNGP) ~\cite{lee2017deep}.

The main contribution of this paper is to extend the concept of the KIP method to a broader range of loss functions. Using the method proposed in this paper, dataset distillation can be performed without bi-level optimization, not only for squared loss but also for commonly used loss functions in classification problems, such as cross-entropy loss and hinge loss. The fundamental idea of the proposed method is to introduce the concept of \emph{Duality Gap (DG)} from the field of mathematical optimization. The DG is a measure that quantifies how close the current solution is to the optimal solution in an optimization problem. Our key idea is to quantify how close the solution based on synthetic instances is to the pre-distillation solution using the DG, enabling dataset distillation to be formulated as a single-level optimization problem. We call the proposed method \emph{Duality Gap KIP (DGKIP)}. Figure~\ref{solutionspace} illustrates the basic idea behind the proposed DGKIP method.

Our contributions are as follows:
\begin{enumerate} 
\item We propose a novel dataset distillation method that avoid bi-level optimization based on DG. Theoretical analysis shows that models trained on the distilled datasets maintain a bounded distance in parameter space compared to those trained on the full dataset. We could also use DG bound (DGB) to bound the prediction accuracy and test error.  
\item 
By using DG as the optimization objective, the DGKIP method extends its applicability to a class of convex loss functions beyond squared loss. Therefore, DGKIP can easily extend to hinge loss, cross entropy, and other classes of loss functions, thereby broadening KIP’s applicability. 
\item Extensive experiments on three benchmarks including MNIST and CIFAR-10, show the effectiveness about test accuracy and time cost of DGKIP. The code for the proposed method and the scripts to reproduce the experiments are provided as supplementary material.
\end{enumerate}

\subsection{Related Works}
Dataset distillation methods aim to generate small synthetic versions of large datasets without sacrificing model accuracy. In this section, we first review dataset distillation methods based on bi-level optimization. Then we survey methods that simplify inner loop constraint. Finally, we discuss how duality gaps can be employed to guide data reduction.

\subsubsection{Dataset Distillation as bi-level optimization}
An early line of research introduced a bi-level optimization framework for dataset distillation \cite{wang2018dataset}. This framework consists of two main components: an inner loop that trains a model on synthetically generated data, and an outer loop that refines this synthetic data by evaluating performance on the original dataset.

This bi-level approach has demonstrated certain efficiency, achieving near-baseline performance with only a fraction of the original dataset. However, when applied to complex datasets, the benefits of the distilled synthetic data often diminish because of computational bottlenecks. In such a bi-level optimization framework, the outer loop depends on the parameters of the inner loop, so updating the synthetic data requires backpropagating through multiple unrolled steps of the inner optimization. This process is similar to backpropagation through time \cite{werbos1990backpropagation} in recurrent neural networks \cite{yu2019review}, where gradients must flow backward over potentially many iterations. As a result, the gradients reaching the outer loop can either explode or vanish, leading to training instabilities \cite{pascanu2013difficulty, vicol2021unbiased}.

\subsubsection{Inner Loop Simplification} 
While bi-level optimization has shown promise, it poses significant computational and stability challenges. Researchers focus on simplifying the inner loop in dataset distillation. An important line of research involves replacing the inner loop with kernel ridge regression (KRR) \citep{nguyen2020dataset, nguyen2021dataset, zhou2022dataset, loo2022efficient}, which often outperforms alternative methods. This is because the loss function for KRR is expressed in a quadratic form, allowing its solution to be derived analytically and thus avoiding bi-level optimization.

The Kernel Inducing Points (KIP) framework \cite{nguyen2020dataset} moved this line of research forward by harnessing expressive kernel functions, particularly neural tangent kernel (NTK) \cite{jacot2018neural}, to better approximate neural network behaviors. Building on this, Frepo \cite{zhou2022dataset} integrated neural feature extraction into ridge regression, achieving state-of-the-art results within the KRR paradigm.

However, KRR methods, especially with NTK, can be computationally expensive, often requiring 
high-cost computation for kernel matrix operations, even for simple kernels. When using NTK, the complexity further increases, making it challenging to scale. Loo et al. \cite{loo2022efficient} addressed this by introducing Random Feature Approximation (RFA), which replaces NTK with an NNGP kernel approximation \cite{lee2017deep}, reducing the computational cost.

\subsubsection{Duality Gap Bound}

The DG bound used in this study is derived from safe screening literature, a method developed for sparse modeling techniques such as Lasso or SVM \citep{ghaoui2010safe}.
Safe screening aims to identify the potential range of optimal model parameters, enabling the elimination of unnecessary features in Lasso \citep{wang2013lasso,ndiaye2017gap} or unnecessary instances in SVM \citep{ogawa2014safe,shibagaki2016simultaneous} before actually solving the optimization problem.
This concept has led to the development of DG-based techniques \citep{ndiaye2015gap}, which have been employed across various scenarios \citep{ndiaye2015gap,nakagawa2016safe,hanada2018efficiently,hanada2023generalized,zhai2020safe,dantas2021expanding}.
Particularly pertinent to this study is the application of DG bounds in distributionally robust coreset selection \citep{hanada2024distributionally,tanaka2025distributionally}.
Our technical contribution in this study lies in adapting the DG bound to enhance the dataset distillation method, KIP, broadening its compatibility with diverse loss functions.

\section{Preliminary}
\label{Preliminary}

In this section, we present the preliminary materials necessary for understanding the proposed DGKIP method.
First, in section \ref{subsec:primal_dual_formulation}, we formulate the primal and dual representations of the model considered in this study and define the duality gap (DG), which serves as the core concept of this research, based on these primal-dual representations.
Then, in section \ref{subsec:dataset_distillation}, we formulate the dataset distillation problem and outline the fundamental idea of the DGKIP method.

\subsection{Primal and Dual Formulation}
\label{subsec:primal_dual_formulation}

The proposed DGKIP method, like the KIP method, is formulated as a kernel function model, in which, by using NTK or NNGP kernels as the kernel function, data distillation for deep learning models becomes feasible.
In this section, to derive the kernel model, we first consider a linear model in a high-dimensional reproducing kernel Hilbert space (RKHS) as the primal problem and then derive the kernel function model as its dual problem.

Most machine learning methods can be viewed as regularized empirical risk minimization (ERM).
Let $\boldsymbol{x}_i \in \mathbb{R}^d, y_i \in \mathbb{R}(i=1,\ldots,n)$ represent the training inputs and their associated labels.
To derive a kernel-based model, let us first consider a linear model in a RKHS ${\cal H}$:
\begin{equation*}
 f(\bm x_i; \bm \theta) = \bm \phi(\bm x_i)^\top \bm \theta
\end{equation*}
where $\bm \phi: \mathbb{R}^d \rightarrow \mathcal{H}$ is a feature mapping function into the RKHS ${\cal H}$, and $\boldsymbol{\theta}$ represents the model parameters.
Given a loss function $\ell(y_i, f(\boldsymbol{x}_i; \boldsymbol{\theta}))$ and a regularization term $\psi(\boldsymbol{\theta})$, we define the optimal parameter $\boldsymbol{\theta}^*$ as
\begin{equation}
\boldsymbol{\theta}^* := \arg \min_{\boldsymbol{\theta} \in \mathbb{R}^d} P(\boldsymbol{\theta}),
\label{eq:primal}
\end{equation}
where
\begin{equation*}
 P(\boldsymbol{\theta}) := \sum_{i=1}^n \ell(y_i, f(\boldsymbol{x}_i;\boldsymbol{\theta})) + \psi(\bm \theta).
\end{equation*}
Here, both $\ell$ and $\psi$ are assumed to be convex and continuous.
In this paper, we focus on $L_2$ regularization term defined as $\psi(\boldsymbol{\theta})=\frac{\lambda}{2}\|\boldsymbol{\boldsymbol{\theta}}\|^2_2$ although the proposed method can go beyond the setting of $L_2$regularization, but it will be more complex to formulate.

We can invoke Fenchel duality to transform this primal problem into a corresponding dual problem. In the dual space, we introduce a dual variable $\alpha_i$ for each training instance.
Let $\ell^*$ and $\psi^*$ be the convex conjugates (Appendix \ref{app:Conjugate}) of $\ell$ and $\psi$, respectively. (For $\ell^*$, the convex conjugate is taken for the second argument of $\ell$).
Then, the dual problem can be written as
\begin{equation}
 \boldsymbol{\alpha}^* := \arg\max_{\boldsymbol{\alpha}\in\mathbb{R}^n} D(\boldsymbol{\alpha}),
  \label{eq:dual}
\end{equation}
where $D(\bm \alpha)$ is the dual objective defined as
\begin{equation*}
 D(\boldsymbol{\alpha}) := -\sum_{i=1}^n \ell^*(y_i, -\alpha_i) - \frac{1}{2\lambda}\sum_{i=1}^n \sum_{i=j}^n\alpha_i\alpha_j y_i y_j k (\boldsymbol{x}_i,\boldsymbol{x}_j).
\end{equation*}
Here, both $\ell$ and $\psi$ are assumed to be convex and continuous, and $k(\boldsymbol{x}_i,\boldsymbol{x}_j)= \bm \phi(\boldsymbol{x}_i)^\top \bm \phi(\boldsymbol{x}_j)$ is a kernel function.
%
%
%
Under certain regularity conditions (e.g., Slater's condition), strong duality holds, meaning the optimal values of the primal and dual objectives coincide:
\begin{equation*}
 P(\boldsymbol{\theta}^*)=D(\boldsymbol{\alpha^*}).
\end{equation*}

Furthermore, the Karush–Kuhn–Tucker (KKT) conditions ensure that, the primal and dual solutions satisfy certain subgradient relationships at the optimum, such as
\begin{equation}
 \boldsymbol{\theta}^* = \frac{1}{\lambda}\sum_{i=1}^n \alpha_i y_i \bm \phi(\boldsymbol{x}_i),\quad-\boldsymbol{\alpha}^* \in \partial\ell(\boldsymbol{y}, f(X ;  \boldsymbol{\theta}^*)),
 \label{eq:ktt1}
\end{equation}
where $\partial \ell(\cdot)$ is the subgradient for the second argument.
Using the dual variables, the classifier $f$ is written as a kernel model in the form of 
\begin{equation*}
 f(\boldsymbol{x};\boldsymbol{\theta})=\sum_{i=1}^n \alpha_i y_i k\left(\boldsymbol{x}_i, \boldsymbol{x}\right)
\end{equation*}
For any primal solution $\boldsymbol{\theta}$ and dual solution $\boldsymbol{\alpha}$, the DG is defined as
\begin{equation}
 {\rm DG}(\bm \theta, \bm \alpha) := P(\boldsymbol{\theta}) - D(\boldsymbol{\alpha}).
\label{eq:duality_gap}
\end{equation}
This gap indicates how close the current solution is to the optimal solution; it reaches zero if and only if $\boldsymbol{\theta} = \boldsymbol{\theta}^*$ and $\boldsymbol{\alpha} = \boldsymbol{\alpha}^*$ are both optimal.

\subsection{Dataset Distillation}
\label{subsec:dataset_distillation}

In this study, we consider dataset distillation for a binary classification problem.
In order to differentiate the datasets before and after dataset distillation, let us denote the original dataset as $\cO = \left(X_{\cO}, \bm y_{\cO}\right) = \left\{(\bm x_i^{\cO}, y_i^{\cO}) \right\}_{i=1}^{n_{\cO}}$ where $n_{\cO}$ is the number of instances, $X_\cO \in \RR^{n_{\cO} \times d}$ is the input matrix, and $\bm y_{\cO} \in \{-1, +1\}^{n_{\cO}}$ is the corresponding label vector for the original dataset.
Given the model with parameter $\bm \theta$, the classification error on the training set is defined as 
\begin{equation*}
 {\rm TrEr}_\cO(\bm \theta)
 =
 \frac{1}{n_\cO}
 \sum_{i=1}^{n_\cO}
 I(y^\cO_i \neq {\rm sgn}\left( f(\bm x^\cO_i; \bm \theta)\right)),
\end{equation*}
where $I(\cdot)$ is the indicator function that returns 1 if the argument is true, while 0 otherwise, and ${\rm sgn}(\cdot)$ is the sign function that returns the sign of the argument.
For a test dataset, which follows the same distribution as the original training dataset, classification error on the test is similarly defined as ${\rm TeEr}_\cO(\bm \theta)$~\footnote{
In this paper, we assume that reducing the training error by appropriately controlling the model capacity during training will also reduce the test error. Therefore, in deriving the methodology, we do not explicitly distinguish between them. However, in the experiments, we use the test error \({\rm TeEr}_\cO(\bm \theta)\) as the evaluation metric.
}.
Therefore, the goal is to find the model parameter $\bm \theta$ that minimize ${\rm TrEr}_{\cO}(\bm \theta)$.

However, in the ordinary classification problem, since it is difficult to minimize the non-differentiable classification error directly, the model parameter $\bm{\theta}$ is trained to minimize the surrogate loss function, such as cross-entropy loss or hinge loss, on the original training dataset, i.e.,
\begin{equation}
 \bm \theta_{\cO}^*
 =
 \argmin_{\bm \theta}
 \sum_{i=1}^{n_{\cO}}
 \ell
 \left(
 y^{\cO}_i, f(\bm x^{\cO}_i; \bm \theta)
 \right)
 +
 \psi\left(\bm \theta\right).
\label{eq:theta_o-opt}
\end{equation}
This is justified because minimizing loss functions tailored to classification problem, such as cross-entropy loss or hinge loss, leads to the minimization of classification error ${\rm TrEr}_{\cO}(\bm \theta)$ and ${\rm TeEr}_{\cO}(\bm \theta)$.

In dataset distillation setting, our goal is to generate a much smaller synthetic dataset denoted as $\cS=\left(X_{\cS}, \bm y_{\cS}\right)=\left\{(\bm x^{\cS}_i, y^{\cS}_i)\right\}_{i=1}^{n_{\cS}}$ such that the number of reduced synthetic dataset size $n_{\cS} \ll n_{\cO}$.
With the reduced synthetic dataset, the model parameter is trained in the same way as
\begin{equation}
 \label{eq:theta_s-opt}
 \bm \theta_{\cS}^*
 =
 \argmin_{\bm \theta}
 \sum_{i=1}^{n_{\cS}}
 \ell
 \left(
 y^{\cS}_i, f(\bm x^{\cS}_i; \bm \theta)
 \right)
 +
 \psi\left(\bm \theta\right),
\end{equation}
Unfortunately, $\bm \theta_\cS^*$ is not guaranteed to lead to the minimization of classification error on the original dataset, i.e., ${\rm TrEr}_\cO(\bm \theta)$ or ${\rm TeEr}_\cO(\bm \theta)$.
Therefore, the goal of dataset distillation is to find $\cS$ such that the $\bm \theta^*_\cS$ minimizes the classification error on the original dataset, which is formally written as 
\begin{equation}
 \label{eq:bi-level-opt}
 \cS^* = \argmin_{\cS} {\rm TrEr}_\cO(\bm \theta^*_\cS).
\end{equation}
Note that the optimization problem in \eq{eq:bi-level-opt} is a bi-level optimization problem because the objective function includes \(\bm{\theta}_\cS^*\), which is the optimal solution of another optimization problem in \eq{eq:theta_s-opt}.
%
%
As mentioned in \S\ref{Introduction}, since directly solving the bi-level optimization problem in \eq{eq:bi-level-opt} is challenging, various approaches have been proposed to approximate its solution.

The basic idea behind the proposed DGKIP method is to find the synthetic dataset $\cS$ that minimize \emph{an upper bound} of the difference between $\bm \theta_\cO^*$ and $\bm \theta_\cS^*$, which can be obtained without actually solving the inner-loop optimization problem in \eq{eq:theta_s-opt}.
This approach is justified in the sense that the minimizing \emph{an upper bound} of the difference between $\bm \theta_\cO^*$ and $\bm \theta_\cS^*$ leads to minimizing an upper bound of ${\rm TrEr}_\cO(\bm \theta)$ and ${\rm TeEr}_\cO(\bm \theta)$.

\subsection{How does KIP generalize to DGKIP?}
KIP considers the problem in \eqref{eq:theta_o-opt} with $L_2$ regularization (KRR) optimized by mean square error (MSE) loss. Under this setting, the optimization problem can be formulated as:
\begin{equation*}
    \min_{\bm \theta_{\cS}}
    \sum_{i=1}^{n_{\cO}}\bigl\|y^{\cO}_i - \bm \phi(x^{\cO}_i)^\top \bm \theta_{\cS} \bigr\|_2^2, 
\end{equation*}
where the optimal solution $\bm \theta_{\cS}^*$ is calculated in closed form as
\begin{equation*}
\bm \theta_{\cS}^* = \bm \phi(X_\mathcal{S}) (\bm \phi(X_\mathcal{S})^\top \bm \phi(X_\mathcal{S}) + \lambda I)^{-1} y_{\mathcal{S}}.
\end{equation*}
This approach then eliminates the need for bi-level optimization.  
By replacing $ y_\mathcal{O} $ with $ \bm \phi(X_\mathcal{O})^\top \boldsymbol{\theta}_\mathcal{O}^* $, we immediately obtain the following equivalence:
\[
\min_{\mathcal{S}} 
\sum_{i=1}^{n_{\cO}}
\bigl\|\boldsymbol{\theta}_\mathcal{O}^* - \boldsymbol{\theta}_\mathcal{S}^* \bigr\|_2^2 \| \bm \phi(x^{\cO}_i)\|^2 \propto \min_{\mathcal{S}} \bigl\|\boldsymbol{\theta}_\mathcal{O}^* - \boldsymbol{\theta}_\mathcal{S}^* \bigr\|_2^2.
\]
This result directly aligns with our formulation, which aims to minimize the discrepancy between model parameters. In section \ref{Methods}, we prove that the upper bound of 
$\bigl\|\bm \theta_{\cO}^* - \bm \theta_{\cS}^* \bigr\|_2^2$ can be optimized by DG in \eqref{eq:duality_gap} and apply to a boarder class of convex loss functions.

\section{Methods}
\label{Methods}
In this section, we explain the proposed DGKIP method in details. First, we propose a theorem to calculate the DG bound and use it to derive the prediction upper bound and the test error upper bound. Then we explain the general pipeline of DGKIP. Finally, we showcase an example of logistic regression formulation.

\subsection{Duality Gap Bound}
\label{duality_gap_bound}

A key assumption of DGKIP is that the regularization term and potentially the entire objective is \(\lambda\)-strongly convex. 
\begin{definition}[\(\lambda\)-strong convexity]
  Let \(\lambda > 0\), and consider a function
  \(f: \mathbb{R}^d \to \mathbb{R} \cup \{\infty\}\) that is convex and has a nonempty domain.
  We say that \(f\) is \(\lambda\)-strongly convex if, for all
  \(\boldsymbol{\theta}, \boldsymbol{\theta}' \in \mathrm{dom}(f)\) and for every
  subgradient \(\boldsymbol{g} \in \partial f(\boldsymbol{\theta}')\), the following holds:
  \[
    f(\boldsymbol{\theta}) - f(\boldsymbol{\theta}') 
    \;\ge\;
    \boldsymbol{g}^\top \!\bigl(\boldsymbol{\theta} - \boldsymbol{\theta}'\bigr) 
    \;+\; \tfrac{\lambda}{2}\,\|\boldsymbol{\theta} - \boldsymbol{\theta}'\|_2^2.
  \]
  \end{definition}


The bound on machine learning model parameters based on the DG was first introduced in the context of safe screening. By adapting the idea of the DG bound for safe screening \cite{shibagaki2016simultaneous} to the dataset distillation problem, we obtain the following theorem.

\begin{theorem}[Bound on Parameter Deviation]
\label{thm:parameter_bound}
Suppose the objective function $P_{\mathcal{O}}(\boldsymbol{\theta})$ and $P_{\mathcal{S}}(\boldsymbol{\theta})$ are $\lambda$-strongly convex with respect to $\boldsymbol{\theta}$, and let $\bm \theta_{\cO}^*$ be its unique minimizer. Let $\bm \theta_{\cS}^*$ be the corresponding minimizer when the dataset is replaced by a smaller synthetic set $\mathcal{S}$. 
The duality gap is defined as
\[
  G_{\mathcal{S}}\bigl(\bm \theta_{\cO}^*, \tilde{\boldsymbol{\alpha}}_{\mathcal{S}}\bigr)
  \;:=\;
  P_{\mathcal{S}}\bigl(\boldsymbol{\theta}_{\cO}^*\bigr)
  \;-\;
D_{\mathcal{S}}\bigl(\tilde{\boldsymbol{\alpha}}_{\mathcal{S}}\bigr),
\]
where
\begin{equation*}
   \begin{aligned}
    P_\mathcal{S}(\bm \theta_{\cO}^*)&:= \sum_{i=1}^{n_\mathcal{S}} \ell(y_i^\mathcal{S}, f (\boldsymbol{x}_i^\mathcal{S};  \bm \theta_{\cO}^*))+\frac{\lambda}{2}\|\bm \theta_{\cO}^*\|^2  \\
D_{\mathcal{S}}\bigl(\tilde{\boldsymbol{\alpha}}_{\mathcal{S}}\bigr)&:= \sum_{i=1}^{n_\mathcal{S}} \ell^*(y_i^\mathcal{S}, -\tilde{\alpha}_i^\mathcal{S})\\
&- \frac{1}{2\lambda}\sum_{i=1}^{n_\mathcal{S}} \sum_{i=j}^{n_\mathcal{S}}\tilde{\alpha}_i^\mathcal{S}\tilde{\alpha}_j^\mathcal{S} y_i^\mathcal{S} y_j^\mathcal{S} k (\boldsymbol{x}_i^\mathcal{S},\boldsymbol{x}_j^\mathcal{S}).
    \end{aligned}
\end{equation*}
Here, $\tilde{\boldsymbol{\alpha}}_{\mathcal{S}} \in \mathrm{dom}(D_{\mathcal{S}})$ is any feasible dual variable. Then, the deviation between the parameter solutions satisfies
\begin{equation}
      \bigl\|\bm \theta_{\cO}^* \;-\; \bm \theta_{\cS}^*\bigr\|_2^2
  \;\;\le\;\;
  \frac{2}{\lambda}\;
  G_{\mathcal{S}}(\bm \theta_{\cO}^*, \tilde{\boldsymbol{\alpha}}_{\mathcal{S}}).
\label{eq:dualitygap} 
\end{equation}

\end{theorem}
The proof of Theorem \ref{thm:parameter_bound} is present in Appendix \ref{proof_theorem}.

The feasible dual variable  $\tilde{\boldsymbol{\alpha}}_\mathcal{S}$ can be constructed arbitrarily as long as it satisfies the constraint conditions. In this work, to tighten the bound, $\tilde{\boldsymbol{\alpha}}_\mathcal{S}$ is set based on the KKT conditions \eqref{eq:ktt1} as follows:
\begin{equation}
  \tilde{\boldsymbol{\alpha}}_{\mathcal{S}} \in -\partial \ell\left(\boldsymbol{y_\mathcal{S}},  f (X_\mathcal{S};  \bm \theta_{\cO}^*)\right).
  \label{eq:alphaapproxi}
\end{equation}
Given the DGB in \eqref{eq:dualitygap}, we could say large DG will lead to loose bounds. In contrast, small DG can provide tight bounds. With a tight bound, we could derive the bound of prediction values. For any input from the training set, if
$\boldsymbol{\theta}_{\mathcal{S}}$ lies within an $L_2$-ball around
$\boldsymbol{\theta}_{\mathcal{O}}$, then for any vector
$\boldsymbol{x} \in \mathbb{R}^d$,
  \begin{equation*}
    \begin{aligned}
      \bigl|f(\boldsymbol{x};\bm \theta_{\cS}^*)\;-\;f(\boldsymbol{x};\bm \theta_{\cO}^*) \bigr| & \;\;\le\;\;\|\boldsymbol{\phi}(\boldsymbol{x})\|_2 \,\cdot\,\bigl\|\bm \theta_{\cO}^* \;-\; \bm \theta_{\cS}^*\bigr\|_2^2.\\                            & \;\;\le\sqrt{k(\boldsymbol{x},\boldsymbol{x}) \,\cdot\, \frac{2}{\lambda}G_{\mathcal{S}}(\bm \theta_{\cO}^*,\tilde{\boldsymbol{\alpha}}_{\mathcal{S}})}.
    \end{aligned}
  \end{equation*}
This immediately yields an upper bound on per-sample (and thus aggregate) prediction deviations after dataset distillation.

\begin{lemma}[Minimizing DG Also Minimizes the Prediction Upper Bound]
\label{lem:dg-minimizes-prediction-bound}
Let $G_{\mathcal{S}}$ be the duality gap defined in Theorem~\ref{thm:parameter_bound}. If $P_{\mathcal{O}}(\boldsymbol{\theta})$ is $\lambda$-strongly convex, then minimizing this duality gap with respect to $\mathcal{S}$ also minimizes the following quantity for any $\boldsymbol{x}$:
\begin{equation*}
  \min_{\boldsymbol{\theta} 
        \,\in\,
        \bigl\{
          \|\boldsymbol{\theta}
              -\bm \theta_{\cO}^*\|_2
          \,\le\,
          \sqrt{\tfrac{2}{\lambda}G_{\mathcal{S}}(\bm \theta_{\cO}^*, \tilde{\boldsymbol{\alpha}}_{\mathcal{S}})}
        \bigr\}}
   \!\!
   \bigl|\boldsymbol{\phi}(\boldsymbol{x})^{\top}\boldsymbol{\theta}
          \;-\;
          \boldsymbol{\phi}(\boldsymbol{x})^{\top}\bm \theta_{\cO}^*\bigr|.
\end{equation*}
Equivalently, shrinking the duality gap, reduces the worst-case difference in predicted values 
$\boldsymbol{\phi}(\boldsymbol{x})^{\top}\boldsymbol{\theta}$ 
compared with 
$\boldsymbol{\phi}(\boldsymbol{x})^{\top}\bm \theta_{\cO}^*$.
\end{lemma}

The proof of Lemma \ref{lem:dg-minimizes-prediction-bound} is presented in Appendix \ref{proof_lemma3.2}

Now we have the prediction bound, the instances could be divided into three classes: certainly correct, certainly wrong, and unknown. When we apply it to the test set, test error bounds could be obtained.

\begin{lemma}
\label{lem:test-error-upper-bound}
The range of the test error (\textrm{TeEr}) on the test dataset $\left\{\left(\boldsymbol{x}_i^{\prime}, y_i^{\prime}\right)\right\}_{i=1}^{n^\prime}$ is derived using the bound of model parameters after retraining as follows:
    \begin{equation*}
\frac{n_{\text{mis}}}{n^\prime} \leq \textrm{TeEr} \leq \frac{n_{\text{mis}} + n_{\text{unk}}}{n^\prime} = \frac{n^\prime - n_{\text{cor}}}{n^\prime}
\end{equation*}
\[
n_{\text{cor}} = \sum_{i=1}^{n^\prime}  I\left[\min_{\boldsymbol{\theta}\in\Theta} y_i^\prime\boldsymbol{\theta}^\top\boldsymbol{\phi}(\boldsymbol{x}_i^\prime) > 0\right],
\]
\[
n_{\text{mis}} =  \sum_{i=1}^{n^\prime}  I\left[\max_{\boldsymbol{\theta}\in\Theta} y_i^\prime\boldsymbol{\theta}^\top\boldsymbol{\phi}(\boldsymbol{x}_i^\prime) < 0\right],
\]
\[
n_{\text{unk}} =\sum_{i=1}^{n^\prime}  I\left[\min_{\boldsymbol{\theta}\in\Theta} y_i^\prime{\bm \theta_{\cO}^*}^\top\boldsymbol{\phi}(\boldsymbol{x}_i^\prime) < 0,\quad \max_{\boldsymbol{\theta}\in\Theta} y_i^\prime{\bm \theta_{\cO}^*}^\top\boldsymbol{\phi}(\boldsymbol{x}_i^\prime) > 0\right],
\]
where
  \begin{equation*}
    \begin{aligned}
     & \min _{\boldsymbol{\theta} \in \Theta} y_i^\prime{\bm \theta_{\cO}^*}^\top\boldsymbol{\phi}(\boldsymbol{x}_i^\prime)\\ =&y_i^\prime{\bm \theta_{\cO}^*}^\top\boldsymbol{\phi}(\boldsymbol{x}_i^\prime)-  \|\boldsymbol{\phi}(\boldsymbol{x^\prime})\|_2 \,\cdot\, \sqrt{\frac{2}{\lambda}G_{\mathcal{S}}({\bm \theta_{\cO}^*},\tilde{\boldsymbol{\alpha}}_{\mathcal{S}})}\\= &y_i^\prime f(\boldsymbol{x}_i^\prime;  {\bm \theta_{\cO}^*})  -  \sqrt{k(\boldsymbol{x}^\prime,\boldsymbol{x}^\prime) \,\cdot\, \frac{2}{\lambda}G_{\mathcal{S}}({\bm \theta_{\cO}^*},\tilde{\boldsymbol{\alpha}}_{\mathcal{S}})}
    \end{aligned}
  \end{equation*}
  \begin{equation*}
    \begin{aligned}
     & \max _{\boldsymbol{\theta} \in \Theta} y_i^\prime{\boldsymbol{\theta}}^\top\boldsymbol{\phi}(\boldsymbol{x}_i^\prime)
     \\= &y_i^\prime f(\boldsymbol{x}_i^\prime;  {\bm \theta_{\cO}^*})  +  \sqrt{k(\boldsymbol{x}^\prime,\boldsymbol{x}^\prime) \,\cdot\, \frac{2}{\lambda}G_{\mathcal{S}}({\bm \theta_{\cO}^*},\tilde{\boldsymbol{\alpha}}_{\mathcal{S}})}
    \end{aligned}
  \end{equation*}

\end{lemma}
The proof related to the inequality shown in Lemma \ref{lem:test-error-upper-bound} is presented in Appendix \ref{proof_lemma3.3}.
We note that the bounds of TeEr can be computed even with the kernel model, that is, ${\bm \theta_{\cO}^*}$ does not need to be explicitly computed. 

In subsequent sections, we use the proposed theorem and lemmas to guide dataset distillation, ensuring that the distilled dataset $\mathcal{S}$ induces a model ${\bm \theta_{\cS}^*}$ that is not excessively distant from the original model ${\bm \theta_{\cS}^*}$.

\subsection{Overview of DGKIP}

DGKIP proceeds through the following key stages in practical dataset distillation. First, a baseline model is trained on the full dataset $\mathcal{O}$ to obtain parameters ${\bm \theta_{\cO}^*}$. This step is performed only once and serves as the reference solution. Next, a smaller synthetic set $\mathcal{S}$ with size $n_\mathcal{S}$ is initialized, either randomly or by a heuristic. DGKIP then uses ${\bm \theta_{\cO}^*}$ as a guide to approximate the dual variables $\tilde{\boldsymbol{\alpha}}$ for $\mathcal{S}$ by \eqref{eq:alphaapproxi}. 

With $\tilde{\boldsymbol{\alpha}}_\mathcal{S}$ and $\mathcal{S}$, we could calculate DG by \eqref{eq:dualitygap}. After each update of $\mathcal{S}$, we will approximate $\tilde{\boldsymbol{\alpha}}_\mathcal{S}$. Because the ${\bm \theta_{\cO}^*}$ is pre-compute, so the computation cost can be ignored. These two processes will iterate until convergence. A detailed description of each optimization step is shown in Algorithm \ref{alg:distillation}.

To efficiently incorporate neural-network-like kernels into the DGKIP, we adopt the NNGP random feature approximation \cite{loo2022efficient} to handle large-scale datasets, details is presented in Appendix \ref{nngp_intro}. In the following section, we will illustrate the DGKIP in a specific example of logistic regression.

\subsection{Duality Gap for Logistic Regression (Binary Cross-entropy Loss)}
We consider a binary classification logistic regression with a $L_2$ regularization term. The corresponding dual objective involves variables $\alpha_i \in [0,1]$ for each training example, which captures how strongly each example influences the decision boundary.
With label set $y_i \in \{-1,+1\}$, we solve the following primal problem:
\begin{equation*}
\label{lr-theta-opt}
  P(\bm \theta) =  \sum_{i=1}^n \log[1+\exp(-y_i f( \boldsymbol{x}_i;\boldsymbol{\theta}))] + \frac{\lambda}{2} \|\boldsymbol{\theta}\|_2^2.
\end{equation*}
Through Fenchel duality, we obtain dual variables $\alpha_i \in [0,1]$ in the following dual problem:
\begin{equation*}
\begin{aligned}
D(\bm \alpha) =  &\frac{1}{2\lambda}\sum_{i=1}^{n}\sum_{j=1}^{n} \alpha_i\alpha_j y_iy_jk(\boldsymbol{x}_i,\boldsymbol{x}_j) \\ & + \sum_{i=1}^{N} \alpha_i \log \alpha_i + (1 - \alpha_i) \log(1 - \alpha_i)
\end{aligned}
\end{equation*}
According to Theorem  \ref{thm:parameter_bound}, the duality gap is formulated as follows 
\begin{equation*}
  \begin{aligned}
G_{\mathcal{S}}(\bm \theta_{\cO}^*,\tilde{\boldsymbol{\alpha}}_\mathcal{S}) = & \sum_{i=1}^{n_{\mathcal{S}}} \log[1+\exp(-y_i^\mathcal{S} f(\boldsymbol{x}_i^\mathcal{S}; \boldsymbol{\theta}_\mathcal{O}))] + \frac{\lambda}{2}\|\boldsymbol{\theta}_\mathcal{O}\|\\
& +\frac{1}{2\lambda}\sum_{i=1}^{n_\mathcal{S}} \sum_{j=1}^{n_\mathcal{S}} \tilde{\alpha}_i^\mathcal{S} \tilde{\alpha}_j^\mathcal{S}  y_i^\mathcal{S} y_j^\mathcal{S} k(\boldsymbol{x}_i^\mathcal{S},\boldsymbol{x}_j^\mathcal{S})\\
& +\sum_{i=1}^{n_{\mathcal{S}}}(\tilde{\alpha}_i^\mathcal{S}\ln\tilde{\alpha}_i^\mathcal{S} + (1-\tilde{\alpha}_i^\mathcal{S})\ln(1-\tilde{\alpha}_i^\mathcal{S})),
  \end{aligned}
\end{equation*}
where
\begin{equation*}
\tilde{\boldsymbol{\alpha}}_{\mathcal{S}} \in -\partial \ell\left(\boldsymbol{y}_\mathcal{S}, f (X_\mathcal{S};  \bm \theta_{\cO}^*)\right)= \sigma(-\boldsymbol{y}_\mathcal{S} f(X_\mathcal{S}; \bm \theta_{\cO}^*)).
\end{equation*}
Here, $\sigma$ is the sigmoid function. Minimizing this gap places the logistic-regression solution on $\mathcal{S}$ close to the corresponding solution on $\mathcal{O}$, thereby preserving classification performance on the original dataset. 

The DGKIP approach avoids bi-level optimization with a single-level optimization on duality gap. By drawing on fundamental results in convex analysis and duality, the proposed framework ensures that shrinking the duality gap brings the distilled solution $\boldsymbol{\theta}_{\mathcal{S}}$ closer to $\boldsymbol{\theta}_{\mathcal{O}}$ in parameter space. This method is broadly applicable to various smooth convex losses. Complete details of the SVM variant are described in Appendix \ref{DGKIP-SVM}.

\begin{algorithm}[t]
\caption{Dataset Distillation Framework}
\label{alg:distillation}
\begin{algorithmic}
    \Require Original dataset $\mathcal{O}$
    \Ensure Distilled dataset $\mathcal{S}$
    \State \textbf{Step 1: Train on } $\mathcal{O}$
        \State \quad Solve primal for LR: 
               ${\bm \theta_{\cO}^*}$ by \eqref{eq:theta_o-opt}
    \State \textbf{Step 2: Initialize } $\mathcal{S}$ \text{ (size $n_\mathcal{S}$)}
    \Repeat
        \State \textbf{Step 3:} Obtain approximate dual $\tilde{\boldsymbol{\alpha}}_\mathcal{S}$ for $\mathcal{S}$
        \State \textbf{Step 4:} Minimize $G_{\mathcal{S}}({\bm \theta_{\cO}^*}, \tilde{\boldsymbol{\alpha}}_\mathcal{S})$ 
                 w.r.t. $\mathcal{S}$
    \Until{convergence}
    \State {\bfseries return} $\mathcal{S}$
\end{algorithmic}
\end{algorithm}
\section{Experiments}
\label{Experiments}

\begin{table*}[p]
\caption{Distillation results on three datasets with varying synthetic dataset sizes. Convolutional neural network random features are extracted for DGKIP-SVM and DGKIP-LR. Bolded numbers indicate the best performance, and underlined numbers indicate the second best performance. All the methods used learned labels. Training times for each method are reported in seconds.}
\label{kernel-distillation-conv}
\begin{center}
\begin{sideways}
\begin{scriptsize}
\begin{sc}
\begin{tabular}{lccccccccccc}
\toprule
Dataset & Img/Cls & Metric & DD & DC & DSA & DGKIP-SVM & DGKIP-LR \\
\midrule
\multirow{6}{*}{MNIST} 
& \multirow{2}{*}{1} & Acc. & 99.56 $\pm$ 0.16 & 98.96$\pm$0.81 & 98.79$\pm$0.99 & \textbf{99.97$\pm$0.02} & \underline{99.75$\pm$0.02} \\
& & Time & 0.237 & 0.186 & 0.192 & \textbf{0.007} & \underline{0.009} \\
\cmidrule{2-8}
& \multirow{2}{*}{10} & Acc. & 99.63 $\pm$ 0.11 & 98.25$\pm$0.56 & \underline{99.77$\pm$0.08} & \textbf{99.99$\pm$0.02} & 99.76$\pm$0.00 \\
& & Time & 0.244 & 0.196 & 0.206 & \underline{0.012} & \textbf{0.011} \\
\cmidrule{2-8}
& \multirow{2}{*}{50} & Acc. & 99.72 $\pm$ 0.11 & 98.96$\pm$0.91 & \underline{99.81$\pm$0.07} & \textbf{99.96$\pm$0.06} & 99.75$\pm$0.02 \\
& & Time & 0.602 & 0.217 & 0.434 & \underline{0.041} & \textbf{0.040} \\
\midrule
\multirow{6}{*}{CIFAR-10}
& \multirow{2}{*}{1} & Acc. & 78.04 $\pm$ 0.06 & 71.58$\pm$1.16 & 69.43$\pm$2.24 & \textbf{85.63$\pm$0.97} & \underline{83.18$\pm$2.68} \\
& & Time & 0.245 & 0.186 & 0.190 & \textbf{0.011} & \textbf{0.011} \\
\cmidrule{2-8}
& \multirow{2}{*}{10} & Acc. & 77.58 $\pm$ 0.02 & 86.47$\pm$0.66 & 82.91$\pm$0.91 & \textbf{90.83$\pm$0.14} & \underline{89.72$\pm$0.10} \\
& & Time & 0.249 & 0.186 & 0.223 & \textbf{0.015} & \textbf{0.015} \\
\cmidrule{2-8}
& \multirow{2}{*}{50} & Acc. & 76.91 $\pm$ 0.02 & \underline{90.45$\pm$0.47} & 88.10$\pm$0.61 & \textbf{91.17$\pm$0.72} & 90.01$\pm$0.02 \\
& & Time & 0.298 & 0.202 & 0.376 & \textbf{0.051} & \textbf{0.051} \\
\midrule
\multirow{6}{*}{Fashion-MNIST}
& \multirow{2}{*}{1} & Acc. & 95.96 $\pm$ 0.06 & 95.13$\pm$0.50 & 96.16$\pm$1.07 & \textbf{97.35$\pm$0.39} & \underline{96.70$\pm$0.71} \\
& & Time & 0.295 & 0.123 & 0.184 & \textbf{0.008} & \underline{0.009} \\
\cmidrule{2-8}
& \multirow{2}{*}{10} & Acc. & 95.72 $\pm$ 0.18 & 98.01$\pm$0.41 & 97.30$\pm$0.36 & \textbf{98.70$\pm$0.15} & \underline{98.32$\pm$0.11} \\
& & Time & 0.353 & 0.139 & 0.213 & \textbf{0.011} & \textbf{0.011} \\
\cmidrule{2-8}
& \multirow{2}{*}{50} & Acc. & 95.99 $\pm$ 0.04 & \textbf{98.82$\pm$0.15} & \underline{98.39$\pm$0.35} & 98.57$\pm$0.25 & 98.43$\pm$0.02 \\
& & Time & 0.597 & 0.203 & 0.415 & \textbf{0.041} & \textbf{0.041} \\
\bottomrule
\end{tabular}
\end{sc}
\end{scriptsize}
\end{sideways}
\end{center}
\end{table*}

 \begin{table*}[p]
\caption{Distillation results on three datasets with varying synthetic dataset sizes. Fully connected neural network random features are extracted for DGKIP-SVM and DGKIP-LR. Bolded numbers indicate the best performance, and underlined numbers indicate the second best performance. All the methods used learned labels. Training times for each method are reported in seconds.}
\label{kernel-distillation-mlp}
\begin{center}
\begin{sideways}
\begin{scriptsize}
\begin{sc}
\begin{tabular}{lccccccccccc}
\toprule
Dataset & Img/Cls & Metric & DD & DC & DSA & DGKIP-SVM & DGKIP-LR \\
\midrule
\multirow{6}{*}{MNIST} 
& \multirow{2}{*}{1} & Acc. & \textbf{99.91$\pm$0.00} & 97.11$\pm$1.55 & 99.60$\pm$0.09 & \underline{99.90$\pm$0.02} & 94.44$\pm$6.95 \\
& & Time & 0.128 & 0.065 & 0.069 & \underline{0.026} & \textbf{0.025} \\
\cmidrule{2-8}
& \multirow{2}{*}{10} & Acc. & \underline{99.86$\pm$0.00} & 98.44$\pm$0.87 & 99.77$\pm$0.09 & \textbf{99.93$\pm$0.02} & 99.65$\pm$0.35 \\
& & Time & 0.132 & 0.067 & 0.151 & \textbf{0.025} & \underline{0.026} \\
\cmidrule{2-8}
& \multirow{2}{*}{50} & Acc. & \underline{99.91$\pm$0.00} & 99.78$\pm$0.11 & 99.89$\pm$0.08 & \textbf{99.92$\pm$0.03} & 99.83$\pm$0.02 \\
& & Time & 0.133 & 0.072 & 0.326 & \textbf{0.026} & \underline{0.028} \\
\midrule
\multirow{6}{*}{CIFAR-10}
& \multirow{2}{*}{1} & Acc. & 75.90$\pm$0.00 & 71.45$\pm$0.83 & 71.73$\pm$0.73 & \underline{80.22$\pm$3.35} & \textbf{83.63$\pm$1.95} \\
& & Time & 0.137 & 0.061 & 0.069 & \underline{0.028} & \textbf{0.026} \\
\cmidrule{2-8}
& \multirow{2}{*}{10} & Acc. & 77.00$\pm$0.00 & 80.05$\pm$0.92 & 80.58$\pm$0.87 & \textbf{88.23$\pm$0.38} & \underline{86.85$\pm$0.46} \\
& & Time & 0.131 & 0.063 & 0.137 & \textbf{0.025} & \underline{0.026} \\
\cmidrule{2-8}
& \multirow{2}{*}{50} & Acc. & 76.40$\pm$0.00 & 77.37$\pm$1.15 & 76.32$\pm$0.98 & \textbf{89.49$\pm$0.27} & \underline{87.53$\pm$0.04} \\
& & Time & 0.137 & 0.065 & 0.328 & \textbf{0.026} & \underline{0.028} \\
\midrule
\multirow{6}{*}{Fashion-MNIST}
& \multirow{2}{*}{1} & Acc. & \textbf{98.25$\pm$0.00} & 88.27$\pm$1.03 & 88.38$\pm$1.07 & \underline{97.77$\pm$0.89} & 89.01$\pm$5.58 \\
& & Time & 0.126 & 0.063 & 0.069 & \underline{0.026} & \textbf{0.025} \\
\cmidrule{2-8}
& \multirow{2}{*}{10} & Acc. & \underline{98.40$\pm$0.00} & 97.96$\pm$0.32 & 97.77$\pm$0.34 & \textbf{98.95$\pm$0.08} & 98.14$\pm$0.08 \\
& & Time & 0.127 & 0.063 & 0.137 & \textbf{0.025} & \underline{0.026} \\
\cmidrule{2-8}
& \multirow{2}{*}{50} & Acc. & 97.95$\pm$0.00 & 98.54$\pm$0.13 & 98.47$\pm$0.16 & \textbf{98.89$\pm$0.12} & \underline{98.63$\pm$0.04} \\
& & Time & 0.131 & 0.064 & 0.348 & \textbf{0.026} & \underline{0.028} \\
\bottomrule
\end{tabular}
\end{sc}
\end{scriptsize}
\end{sideways}
\end{center}
\end{table*}

\begin{figure*}[ht]
\vskip 0.2in
\begin{center}
    \begin{minipage}{0.32\textwidth}
        \centerline{\includegraphics[width=\columnwidth]{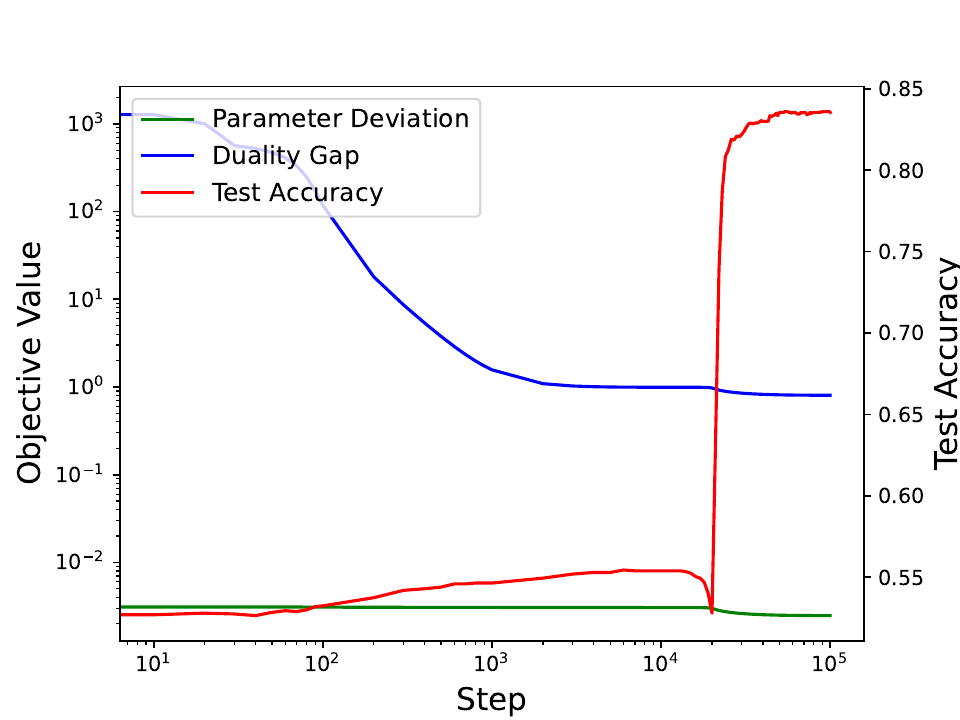}}
        \centerline{\raisebox{1pt}{(a) IPC=1}}
    \end{minipage}
    \hfill
    \begin{minipage}{0.32\textwidth}
        \centerline{\includegraphics[width=\columnwidth]{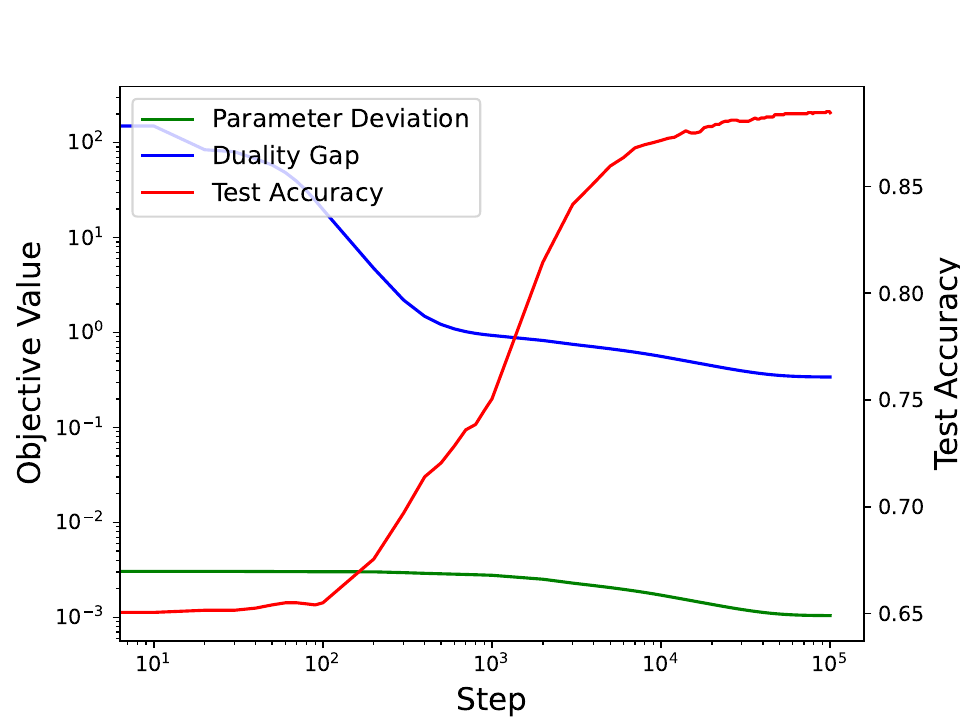}}
        \centerline{\raisebox{1pt}{(b) IPC=10}}
    \end{minipage}
    \hfill
    \begin{minipage}{0.32\textwidth}
        \centerline{\includegraphics[width=\columnwidth]{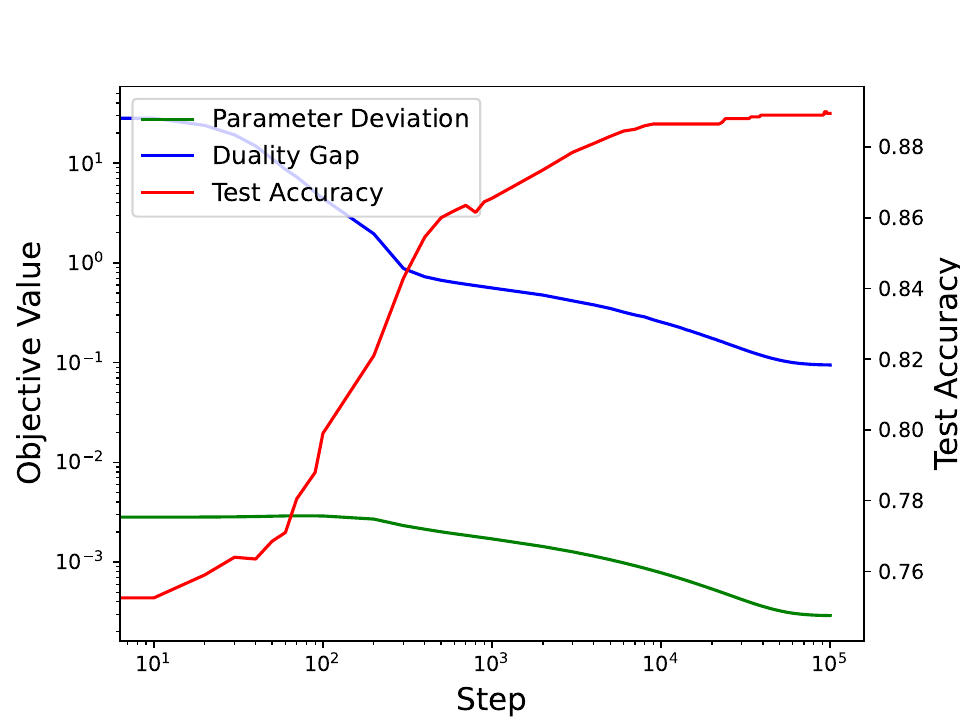}}
        \centerline{\raisebox{1pt}{(c) IPC=50}}
    \end{minipage}
\caption{Parameter deviation, duality gap, and test accuracy varying cross training process with 1, 10, 50 IPC on CIFAR10. The parameter deviation in the green line (left-hand side of Eq.~\eqref{eq:dualitygap}) and the duality gap in the blue line (right-hand side of Eq.~\eqref{eq:dualitygap}) show the same pattern. Minimizing the duality gap reduces the parameter deviation, leading to an increase in test accuracy.}
\label{learning_process}
\end{center}
\end{figure*}

In this section, we first demonstrate the performance of DGKIP on benchmark datasets and compare it with existing data distillation methods. We analyze the method's performance in terms of test accuracy and speed. We also conducted an ablation study on the model used for random feature approximation and the transfer ability when the model used in training is different from the model used in testing. Details setting is presented in Appendix \ref{exp_setting}.

\subsection{Quantitative Results}
To apply the DGKIP method to deep learning models,
we kernelize the method and employ an NNGP kernel equivalent to the target deep learning model. Specifically, We compared our method against established baseline approaches that employed cross-entropy, including DD \cite{wang2018dataset}, DC \cite{zhao2020dataset}, and DSA\cite{zhao2021dataset}. 

\paragraph{Benchmarks.} We applied our algorithm to three datasets: MNIST, Fashion-MNIST and CIFAR-10 \cite{krizhevsky2009learning, lecun2010mnist, xiao2017fashion}, distilling the datasets to synthetic datasets with 1, 10 or 50 images per class (IPC).

Table \ref{kernel-distillation-conv} and Table \ref{kernel-distillation-mlp} present comparative results on those benchmarks. We evaluate the DGKIP implemented in Support Vector Machine (SVM) and Logistic Regression (LR), named DGKIP-SVM and DGKIP-LR. The model used to extract random features are fully connected network (FCNet) and convolutional nerual network (ConvNet), details are also presented in Appendix \ref{exp_setting}. We conducted binary classification experiments based on classes 0 and 1. The test accuracy and training time of every iteration are reported. While cross entropy based approaches (DD, DSA, and DC) show substantial increases in training time as IPC grows, both DGKIP variants maintain remarkably low computational overhead throughout. 

Figure \ref{learning_process} illustrates the learning process of DGKIP based using ConvNet with different numbers of IPC, showing the relationship between parameter deviation $\bigl\|\boldsymbol{\theta}_{\mathcal{O}}-\boldsymbol{\theta}_{\mathcal{S}}\bigr\|_2/\bigl\|\boldsymbol{\theta}_{\mathcal{O}}\bigl\|_2$, its upper bound represented by the duality gap$\sqrt{(2 / \lambda) G_{\mathcal{S}}(\boldsymbol{\theta}_{\mathcal{O}}, \tilde{\boldsymbol{\alpha}}_{\mathcal{S}})}/\|\boldsymbol{\theta}_{\mathcal{O}}\bigl\|_2$, and test accuracy. As training progresses, we observe that the duality gap (blue line) consistently decreases, leading to a reduction in the parameter deviation (green line), which ultimately results in improved test accuracy (red line). While test accuracy does not increase steadily when IPC is low (1 IPC), the optimization of the duality gap  remains efficient across all settings (1, 10, and 50 IPC). The results indicate that minimizing the duality gap effectively guides the distillation process toward better model performance, particularly with reasonable IPC values.

Although the basic SVM model restricts the DGKIP-SVM framework to binary classification tasks, we ensured fairness in experimental comparisons by adopting the same binary classification setup when evaluating other methods. Also, the base model for training are set to the same one, e.g. FCNet, ConvNet.

It is important to emphasize that the application scope of the DGKIP method itself is not limited to binary classification. When DGKIP is combined with models supporting multiclass classification, it can also achieve multiclass functionality. This is because DGKIP is fundamentally a generalized approach for Kernel Inducing Points, and its specific applicability depends on the chosen convex method.


\begin{table}[t]
\caption{Transfer experiment results (Algorithm before/after $\rightarrow$ indicates distillation/evaluation)}
\label{transfertest}
\begin{center}
\begin{tabular}{lccc}
\toprule
Dataset & Img/Cls & SVM$\to$LR & LR$\to$SVM \\
\midrule
MNIST & 1 & 99.97±0.02 & 99.76±0.00 \\
& 10 & 99.98±0.02 & 99.76±0.00 \\
& 50 & 99.93±0.02 & 99.78±0.04 \\
\midrule
F-MNIST & 1 & 97.30±0.39 & 96.72±0.69 \\
& 10 & 98.63±0.18 & 98.39±0.07 \\
& 50 & 98.36±0.11 & 98.58±0.08 \\
\midrule
CIFAR-10 & 1 & 85.59±1.04 & 83.23±2.60 \\
& 10 & 85.06±1.52 & 89.73±0.18 \\
& 50 & 88.21±1.09 & 90.41±0.11 \\
\bottomrule
\end{tabular}
\end{center}
\end{table}

\subsection{Transfer Experiments}

To evaluate whether DGKIP is robust to model changes, we conduct transfer experiments where the model used for training is different from testing. The result are shown in Table \ref{transfertest}. The findings show that DGKIP-SVM-distilled data can be successfully utilized by DGKIP-LR, retaining nearly the same accuracy as when evaluated with SVM. Note that we used ConvNet in both models. A similar outcome is observed when DGKIP-LR-distilled data is used to train SVM, with no significant loss in performance. These results suggest that the distilled datasets preserve essential information across different model types, demonstrating consistent cross-model transferability.

\subsection{Discussion}

DGKIP optimizes the distilled dataset by fast compute the duality gap (DG), which measures the distance between the solution fitted to the synthetic data and the solution fitted to the original data. Intuitively, larger DG values correspond to greater changes in the new dataset's features and instance number, potentially introducing instability during convergence. This is obvious when the number of IPCs is set to differ greatly (e.g. 1 vs 50), as this can lead to variations in convergence speed, sometimes exceeding a factor of two. Furthermore, DGKIP is closely tied to the choice of convex loss function. If the loss functions used during training and testing are different, it can result in a performance drop.
\section{Conclusion}
\label{Conclusion}

We have presented Duality Gap KIP (DGKIP), a novel data distillation framework grounded in duality theory. By using the duality gap as an optimization objective, our approach not only avoids bi-level optimization but also generalizes the KIP method to a broader class of convex loss functions, such as hinge loss and cross-entropy. We provide a theoretical proof that minimizing the duality gap is equivalent to minimizing model parameters in solution space and further establishes upper bounds on prediction and test errors.

We construct two variants, DGKIP-SVM and DGKIP-LR, and compare them with existing dataset distillation methods on benchmark datasets. Experimental results show the effectiveness of DGKIP in terms of both classification accuracy and optimization speed. Additionally, we examine its transferability and discuss the impact of changes in the number of images generated per class and the choice of the NNGP kernel.

\section*{Impact Statement}

This work presents a new method of dataset distillation, aiming to reduce storage requirements and computational costs in machine learning while maintaining model performance. The positive societal impacts include privacy advantages by minimizing disclosure of sensitive information and changing the environmental consideration of training large models by reducing the amount of data needed to achieve acceptable accuracy. Currently, we do not identify any immediate, critical ethical concerns specific to our method. 
\section*{Acknowledgments}
This work was partially supported by MEXT KAKENHI (JP20H00601, JP23K16943, JP24K15080), JST CREST (JPMJCR21D3 including AIP challenge program, JPMJCR22N2), JST Moonshot R\&D (JPMJMS2033-05), JST AIP Acceleration Research (JPMJCR21U2), JST ACT-X Grant Number JPMJAX24C3, NEDO (JPNP20006) and RIKEN Center for Advanced Intelligence Project.


\bibliographystyle{plain}
\bibliography{ref}

\appendix
\newpage
\appendix

\onecolumn
\section{Fenchel Duality Theory}
\label{App:StrongConvex}
\subsection{$\lambda$-Strong Convexity and $\nu$-Smoothness}
A convex function $f:R^n \to R\cup\{+\infty\}$ is called $\lambda$-strongly convex ($\lambda>0$) if, for any $\boldsymbol{x},\boldsymbol{y}$ in its domain and any subgradient $\boldsymbol{g}\in\partial f(\boldsymbol{y})$,
\begin{equation*}
f(\boldsymbol{x}) - f(\boldsymbol{y}) \geq \boldsymbol{g}^\top(\boldsymbol{x}-\boldsymbol{y}) + \frac{\lambda}{2} \|\boldsymbol{x}-\boldsymbol{y}\|_2^2.
\label{eq:lambdaconvex}
\end{equation*}
If $f$ is differentiable, this condition is equivalent to $\nabla^2f(\boldsymbol{x})\succeq\lambda I$, i.e., every eigenvalue of the Hessian is at least $\lambda$. Intuitively, $\lambda$-strong convexity enforces a lower bound on the curvature of $f$, guaranteeing that its minimizer is unique and that gradient-based methods can converge more quickly.

A differentiable function $f$ is called $\nu$-smooth ($\nu>0$) if its gradient is $\nu$-Lipschitz continuous:
\begin{equation*}
\|\nabla f(\boldsymbol{x})-\nabla f(\boldsymbol{y})\|_2 \leq \nu \|\boldsymbol{x}-\boldsymbol{y}\|_2, \quad \forall\,\boldsymbol{x},\boldsymbol{y}.
\label{eq:nusmooth}
\end{equation*}

Equivalently, if $f$ is twice differentiable, $\nabla^2f(\boldsymbol{x})\preceq\nu I$ means the Hessian's eigenvalues are all at most $\nu$. Geometrically, $\nu$-smoothness caps how sharply $f$ can bend, preventing its gradient from changing too abruptly.

\subsection{Dual (Conjugate) Relationship}
\label{app:Conjugate}
For a proper, closed, convex function $f$, the conjugate $f^*$ is defined by
\begin{equation*}
f^*(\boldsymbol{y}) = \sup_{\boldsymbol{x}\in\text{dom}(f)}(\langle \boldsymbol{x},\boldsymbol{y}\rangle - f(\boldsymbol{x})).
\label{eq:convexconjugate}
\end{equation*}

It is a well-known result in convex analysis that:
\begin{itemize}
\item If $f$ is $\nu$-smooth, then $f^*$ is $\frac{1}{\nu}$-strongly convex.
\item If $f$ is $\lambda$-strongly convex, then $f^*$ is $\frac{1}{\lambda}$-smooth.
\end{itemize}

Hence, strong convexity in the primal translates to smoothness in the dual, and vice versa. This duality underpins many optimization algorithms that exploit both primal and dual formulations (e.g., proximal methods, mirror descent).
\section{Proof}
\subsection{Derivation of  Duality Gap Bound}
\label{proof_theorem}
Let $P_\mathcal{S}$ be a $\lambda$ strongly convex primal problem for $\mathcal{S}$, and let $D_\mathcal{S}$ be its corresponding dual problem.
Given the optimal solution $\boldsymbol{\theta}_\mathcal{O}^* \in P_\mathcal{S}$ for the original data $\mathcal{O}$ and an arbitrary feasible solution $\tilde{\boldsymbol{\alpha}}_{\mathcal{S}} \in \mathrm{dom}(D_{\mathcal{S}})$ , the following relationship holds.
 \begin{equation}
    \begin{aligned}
\left\|\bm \theta^*_\cO -\bm \theta^*_\cS \right\|_2 \leq & \sqrt{\frac{2\left(P_{\mathcal{S}}\left(\bm \theta^*_\cO \right)-D_{\mathcal{S}}(\tilde{\boldsymbol{\alpha}}_{\mathcal{S}})\right)} {\lambda}}
    \end{aligned}
  \end{equation}
\begin{proof}
We follow the calculation in \cite{ndiaye2015gap,shibagaki2016simultaneous}. 
Given the relation in \eqref{eq:lambdaconvex} in $\lambda$-Strong Convexity, we can have
 \begin{equation}
    \begin{aligned}
      \frac{\lambda}{2}\left\|\bm \theta^*_\cO-\bm \theta^*_\cS\right\|_2^2  & \leq P_{\mathcal{S}}\left(\bm \theta^*_\cO \right)- P_{\mathcal{S}}\left(\bm \theta^*_\cS \right)
       +\partial P_{\mathcal{S}}\left(\bm \theta^*_\cS \right)^{\top}\left(\bm \theta^*_\cS -\bm \theta^*_\cO\right) \\
    &=P_{\mathcal{S}}\left(\bm \theta^*_\cO \right)-P_{\mathcal{S}}\left(\bm \theta^*_\cS\right)\\
    & =P_{\mathcal{S}}\left(\bm \theta^*_\cO \right)-D_{\mathcal{S}}\left(\boldsymbol{\alpha}^*_{\mathcal{S}}\right)\\
    & \leq P_{\mathcal{S}}\left(\bm \theta^*_\cO\right)-D_{\mathcal{S}}(\tilde{\boldsymbol{\alpha}}_{\mathcal{S}}) \\
\Leftrightarrow\left\|\bm \theta^*_\cO -\bm \theta^*_\cS \right\|_2 \leq & \sqrt{\frac{2\left(P_{\mathcal{S}}\left(\bm \theta^*_\cO \right)-D_{\mathcal{S}}(\tilde{\boldsymbol{\alpha}}_{\mathcal{S}})\right)} {\lambda}}
    \end{aligned}
  \end{equation}
\end{proof}



\subsection{Prediction Bound}
\label{proof_lemma3.2}
From Theorem~\ref{thm:parameter_bound}, 
$\|\bm \theta^*_\cS - \bm \theta^*_\cO\|_2^2 
 \,\le\,
 \tfrac{2}{\lambda} 
 G_{\mathcal{S}}\bigl(\bm \theta^*_\cO, \tilde{\boldsymbol{\alpha}}_{\mathcal{S}}\bigr)$.
Thus, as $G_{\mathcal{S}}\bigl(\bm \theta^*_\cO, \tilde{\boldsymbol{\alpha}}_{\mathcal{S}}\bigr)$ decreases, the radius of the permissible ball around $\bm \theta^*_\cO$ likewise decreases. Consequently, the term 
$\bigl|\phi(\boldsymbol{x})^{\top}\bm \theta^*_\cS -\phi(\boldsymbol{x})^{\top}\bm \theta^*_\cO\bigr|$
is bounded above by 
$\|\phi(\boldsymbol{x})\|_2 \sqrt{\tfrac{2}{\lambda}G_{\mathcal{S}}\bigl(\bm \theta^*_\cO, \tilde{\boldsymbol{\alpha}}_{\mathcal{S}}\bigr)}$.
Minimizing $G_{\mathcal{S}}\bigl(\bm \theta^*_\cO, \tilde{\boldsymbol{\alpha}}_{\mathcal{S}}\bigr)$ 
directly minimizes this latter expression, thereby reducing the maximum possible deviation in $\phi(\boldsymbol{x})^{\top}\bm \theta^*_\cS$ for all $\boldsymbol{x}$.

\subsection{Inequality in Test Error Bound}
\label{proof_lemma3.3}
  For any vector $\boldsymbol{a}, \boldsymbol{c} \in \mathbb{R}^n$ and $S>0$,
  \begin{equation*}
    \min _{\boldsymbol{v} \in \mathbb{R}^n:\|\boldsymbol{v}-\boldsymbol{c}\|_2 \leq S} \boldsymbol{a}^{\top} \boldsymbol{v}=\boldsymbol{a}^{\top} \boldsymbol{c}-S\|\boldsymbol{a}\|_2, \quad \max _{\boldsymbol{v} \in \mathbb{R}^n:\|\boldsymbol{v}-\boldsymbol{c}\|_2 \leq S} \boldsymbol{a}^{\top} \boldsymbol{v}=\boldsymbol{a}^{\top} \boldsymbol{c}+S\|\boldsymbol{a}\|_2
  \end{equation*}
  \begin{proof}
      By Cauchy-Schwarz inequality,
  \begin{equation*}
    -\|\boldsymbol{a}\|_2\|\boldsymbol{v}-\boldsymbol{c}\|_2 \leq \boldsymbol{a}^{\top}(\boldsymbol{v}-\boldsymbol{c}) \leq\|\boldsymbol{a}\|_2\|\boldsymbol{v}-\boldsymbol{c}\|_2
  \end{equation*}
  The first inequality holds as equality when $\exists \omega > 0 : \boldsymbol{a} = -\omega(\boldsymbol{v} - \boldsymbol{c})$, and the second inequality holds as equality when $\exists \omega^{\prime} > 0 : \boldsymbol{a} = \omega^{\prime}(\boldsymbol{v} - \boldsymbol{c})$. Furthermore, since $\|\boldsymbol{v} - \boldsymbol{c}\|_2 \leq S$, the following inequality holds:
  \begin{equation*}
    -S\|\boldsymbol{a}\|_2 \leq \boldsymbol{a}^{\top}(\boldsymbol{v}-\boldsymbol{c}) \leq S\|\boldsymbol{a}\|_2
  \end{equation*}
  Equality holds when $\|\boldsymbol{v} - \boldsymbol{c}\|_2 = S$. Additionally, consider $\boldsymbol{v}$ that satisfies the equality conditions of the Cauchy-Schwarz inequality for both cases above:
  \begin{itemize}
    \item For the first inequality to hold as equality: $\boldsymbol{v} = \boldsymbol{c} - \left(\frac{S}{\|\boldsymbol{a}\|_2}\right) \boldsymbol{a}$,
    \item  For the second inequality to hold as equality: $\boldsymbol{v} = \boldsymbol{c} + \left(\frac{S}{\|\boldsymbol{a}\|_2}\right) \boldsymbol{a}$.
  \end{itemize}
\end{proof}
\section{DGKIP in Support Vector Machine formulation}
\label{DGKIP-SVM}
Support Vector Machine (SVM) seeks an optimal hyperplane that maximizes the margin between classes. Beyond linear separability, the kernel trick allows handling of non-linear classification by mapping samples into higher-dimensional feature spaces.

\paragraph{Primal Form}
For a training set $\{(\boldsymbol{x}_i,y_i)\}_{i=1}^n$ with $y_i \in \{-1,+1\}$, the primal objective with $\lambda$-regularization is:
\begin{equation*}
P(\boldsymbol{\theta})=\sum_{i=1}^n \max\{0, 1-y_i f(\boldsymbol{x}_i; \boldsymbol{\theta})\} + \frac{\lambda}{2} \|\boldsymbol{\theta}\|_2^2.
\end{equation*}

\paragraph{Dual Form}
Introducing dual variables $\{\alpha_i\}_{i=1}^n$ with constraints $0 \leq\alpha_i \leq 1$, the dual problem becomes:
\begin{equation*}
  D (\boldsymbol{\alpha})=  \sum_{i=1}^n \alpha_i - \frac{1}{2\lambda}\sum_{i=1}^n\sum_{j=1}^n \alpha_i \alpha_j y_i y_j k(\boldsymbol{x}_i,\boldsymbol{x}_j). 
\end{equation*}

\paragraph{Duality Gap for SVM}
\label{App:DGKIP-SVM}

For the optimal solution $\bm \theta^*_\cO$ of $\mathcal{O}$ and any feasible dual variable $\tilde{\boldsymbol{\alpha}}_{\mathcal{S}}$, the duality gap is defined as follows:
\begin{equation*}
  \begin{aligned}
    G_{\mathcal{S}}(\bm \theta^*_\cO,\tilde{\boldsymbol{\alpha}}_\mathcal{S}) & = \sum_{i=1}^{n_{\mathcal{S}}}\left[\max\{0, 1-y_i^\mathcal{S}f(\boldsymbol{x}_i^\mathcal{S}; \bm \theta^*_\cO)\} -\tilde{\alpha}_i^\mathcal{S}\right]+\frac{\lambda}{2}\|\bm \theta^*_\cO\|_2^2 \\
 &+  \frac{1}{2\lambda}\sum_{i=1}^{n_\mathcal{S}} \sum_{j=1}^{n_\mathcal{S}} \tilde{\alpha}_i^\mathcal{S} \tilde{\alpha}_j^\mathcal{S} y_i^\mathcal{S} y_j^\mathcal{S} k(\boldsymbol{x}_i^\mathcal{S},\boldsymbol{x}_j^\mathcal{S})
  \end{aligned}
\end{equation*}



For SVM, different from LR, updating synthetic data $\mathcal{S}$ (Step 4 of Algorithm \ref{alg:distillation}) becomes difficult if $\tilde{\alpha}_i^{\mathcal{S}}$ is calculated naively. In fact, from equation \eqref{eq:alphaapproxi}, we have
\begin{equation*}
\tilde{\alpha}_i^{\mathcal{S}}\in 
- \partial\ell(y_i^\mathcal{S}, f(\boldsymbol{x}_i^\mathcal{S} ;  \boldsymbol{\theta}^*_\mathcal{O}))=\left\{\begin{array}{cc}
1 & \text{if }\quad y_i^\mathcal{S}f(\boldsymbol{x}_i^\mathcal{S} ;  \boldsymbol{\theta}^*_\mathcal{O})<1 \\
{[0,1]} & \text{if }\quad y_i^\mathcal{S}f(\boldsymbol{x}_i^\mathcal{S} ;  \boldsymbol{\theta}^*_\mathcal{O})=1 \\
0 & \text{if }\quad y_i^\mathcal{S}f(\boldsymbol{x}_i^\mathcal{S} ;  \boldsymbol{\theta}^*_\mathcal{O}) > 1
\end{array}\right.
\end{equation*}
As a result, the gradient with respect to $\boldsymbol{x}_i^\mathcal{S}$and $y_i^\mathcal{S}$ become zero.
Thus, in our implementation, we approximate it by using the smooth sigmoid function $\sigma$
\begin{equation*}
\tilde{\alpha}_i^{\mathcal{S}} \approx \sigma(1-y_i^\mathcal{S}f(\boldsymbol{x}_i^\mathcal{S} ;  \boldsymbol{\theta}^*))
\end{equation*}


\section{Details of Experiments}
\subsection{NNGP Random Feature Approximation}
\label{nngp_intro}
In brief, instead of directly computing the NNGP kernel
\[k^{\text{NNGP}}(\boldsymbol{x}, \boldsymbol{x}') = \mathbb{E}_{\boldsymbol{w}}[{f_{\boldsymbol{w}}(\boldsymbol{x})}^\top f_{\boldsymbol{w}}(\boldsymbol{x}')],\]
we draw a large number $M$ of random weight vectors $\{\boldsymbol{w}_m\}_{m=1}^M$ (e.g., i.i.d. from $\mathcal{N}(0, \sigma_{\boldsymbol{w}}^2I)$) and define the following explicit feature map into a finite-dimensional space:
\[\phi^{\text{NNGP}}({\boldsymbol{x}}) := \frac{1}{\sqrt{M}}[f_{\boldsymbol{w}_1}, f_{\boldsymbol{w}_2}, \ldots, f_{\boldsymbol{w}_M}]^\top.\]
Then, for any pair of samples $\boldsymbol{x}, \boldsymbol{x}' \in \mathbb{R}^d$, the inner product of these features approximates the NNGP kernel:
\[\phi^{\text{NNGP}}(\boldsymbol{x})^\top \phi^{\text{NNGP}}(\boldsymbol{x}') \approx k^{\text{NNGP}}(\boldsymbol{x}, \boldsymbol{x}'). \]
By substituting $\phi^{\text{NNGP}}(\cdot)$ for $\phi(\cdot)$, we obtain a random-feature-based formulation whose solution converges to that of the infinite-width NNGP model as $M \to \infty$.

\subsection{Settings}
\label{exp_setting}
We run all experiments on a single NVIDIA RTX A6000 GPU.
The regularization parameter \(\lambda\) was set according to the number of samples. Specifically, when obtaining the original data solution $\bm \theta^*_\cO$, we set $\lambda = n_\mathcal{O} \times 10^{-6}$, and dataset distillation process, we set $\lambda = n_\mathcal{S} \times 10^{-6}$.
We used Adabelief optimizer \cite{zhuang2020adabelief} with a learning rate of $1e-2$ and $\epsilon=1e-16$.

For random feature approximation, we chose the fully connected network and convolutional network. The fully connected network has three hidden layers with 1024 neurons, each layer initialized with Gaussian-distributed weights. Similarly, the convolutional network has three convolutional layers with 256 convolutional channels per layer initialized with Gaussian Gaussian-distributed weights. In practice, we construct 30 networks for a fully connected network and 8 networks for a convolutional network, concatenating their outputs to obtain a richer embedding.

\end{document}